\documentclass[letterpaper, 10 pt, conference]{ieeeconf}  

\IEEEoverridecommandlockouts                              

\usepackage{amsmath,amsfonts}
\usepackage{algorithmic}
\usepackage{array}
\usepackage[caption=false,font=normalsize,labelfont=sf,textfont=sf]{subfig}
\usepackage{textcomp}
\usepackage{stfloats}
\usepackage{url}
\usepackage{verbatim}
\usepackage{graphicx} 
\usepackage{epstopdf}
\usepackage{amssymb}
\usepackage{booktabs}
\usepackage{multirow}
\usepackage{multicol}
\usepackage{hyperref}
\usepackage{colortbl}
\definecolor{mygray}{gray}{.85}
\def\BibTeX{{\rm B\kern-.05em{\sc i\kern-.025em b}\kern-.08em
    T\kern-.1667em\lower.7ex\hbox{E}\kern-.125emX}}
\usepackage{balance}

\usepackage{etoolbox}
\makeatletter
\patchcmd{\@makecaption}
  {\scshape}
  {}
  {}
  {}
\makeatother
\title{\LARGE \bf
Anyview: General Indoor 3D Object Detection with Variable Frames
}

\author{Zhenyu Wu$^{1}$, Xiuwei Xu$^{2}$, Ziwei Wang$^{3}$, Chong Xia$^{2}$, Linqing Zhao$^{2}$, Jiwen Lu$^{2}$ and Haibin Yan$^{*1}$
\thanks{*Corresponding author.}
\thanks{$^{1}$Zhenyu Wu and Haibin Yan are with the School of Intelligent Engineering and
Automation, Beijing University of Posts and Telecommunications. {\tt\small \{wuzhenyu, eyanhaibin\}@bupt.edu.cn.}}
\thanks{$^{2}$Xiuwei Xu, Linqing Zhao, Chong Xia, and Jiwen Lu are with the Department of Automation, Tsinghua University.  {\tt\small \{xxw21, zhaolinqing, xiac20\}@mails.tsinghua.edu.cn.} {\tt\small lujiwen@tsinghua.edu.cn.}}
\thanks{$^{3}$Ziwei Wang is with the School of Electrical and Electronic Engineering, Nanyang Technological University. {\tt\small \{ziwei.wang\}@ntu.edu.sg}}
}

\begin{document}

\maketitle
\thispagestyle{empty}
\pagestyle{empty}

\begin{abstract}

In this paper, we propose a novel network framework for indoor 3D object detection to handle variable input frame numbers in practical scenarios.
Existing methods only consider fixed frames of input data for a single detector, such as monocular RGB-D images or point clouds reconstructed from dense multi-view RGB-D images.
While in practical application scenes such as robot navigation and manipulation, the raw input to the 3D detectors is the RGB-D images with variable frame numbers instead of the reconstructed scene point cloud.
However, the previous approaches can only handle fixed frame input data and have poor performance with variable frame input.
In order to facilitate 3D object detection methods suitable for practical tasks,
we present a novel 3D detection framework named AnyView for our practical applications, which generalizes well across different numbers of input frames with a single model.
To be specific, we propose a geometric learner to mine the local geometric features of each input RGB-D image frame and implement local-global feature interaction through a designed spatial mixture module.
Meanwhile, we further utilize a dynamic token strategy to adaptively adjust the number of extracted features for each frame, which ensures consistent global feature density and further enhances the generalization after fusion.
Extensive experiments on the ScanNet dataset show our method achieves both great generalizability and high detection accuracy with a simple and clean architecture containing a similar amount of parameters with the baselines.

\end{abstract}

\section{INTRODUCTION}

3D object detection is a fundamental scene understanding problem for robot manipulation~\cite{wu2025momanipvla},  navigation~\cite{yin2024sg}, which aims to detect the 3D bounding boxes and semantic labels from point clouds or images.
Due to the different types of sensors used in different application scenarios, 3D object detection methods usually vary a lot for indoor~\cite{xu20233d} and outdoor~\cite{xia2023lightweight} scenes. We focus on indoor 3D object detection, where the mainstream sensor is RGB-D cameras and the scenes are crowded with objects of multiple categories and sizes.

\begin{figure}[t]
    \centering
    \includegraphics[width=1.0\linewidth]{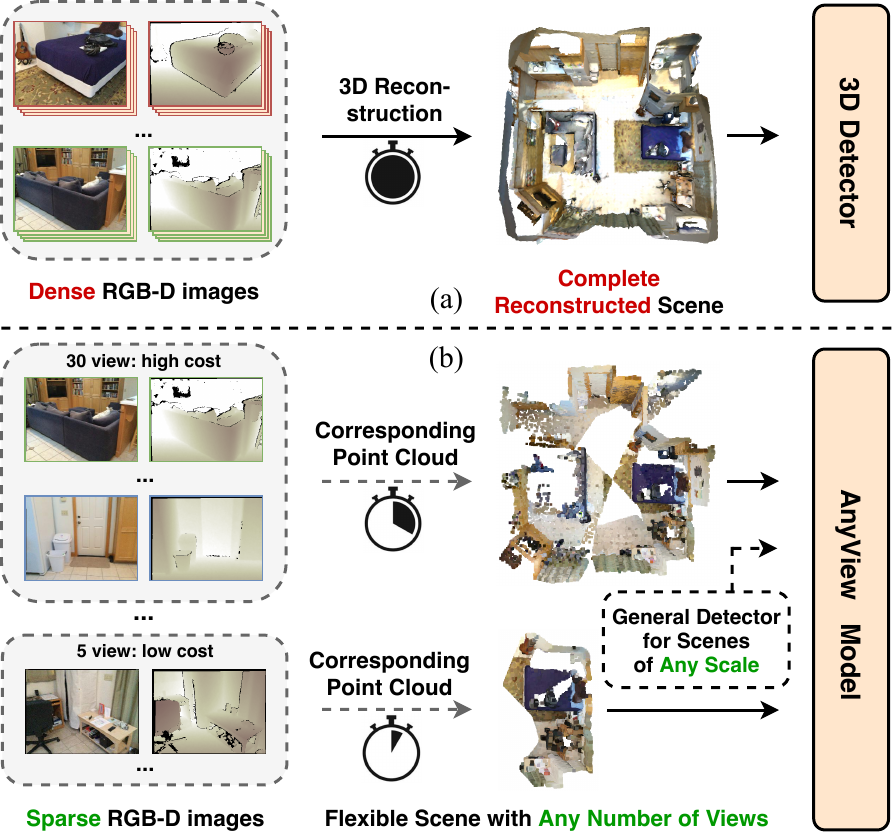}
    \vspace{-7mm}
    \caption{(a) Previous 3D object detection methods rely on reconstructed point clouds from densely sampled RGB-D images, which requires a large amount of time to process the data and generalizes poorly to scenes in different scales. (b) Our method directly detects objects from sparse RGB-D images of any number of views, which shows good generalizability and provides more trade-off.}
    \vspace{-6mm}
    \label{teaser}
\end{figure}

Although great improvement in performance has been achieved by advanced architecture design, existing methods train and evaluate 3D detectors on only fixed frames of input data, such as monocular RGB-D images (on SUN-RGBD~\cite{song2015sun} benchmark) and scene-level reconstructed point clouds (sampled from meshes) from multi-view RGB-D images (on ScanNet~\cite{dai2017scannet} benchmark).
There is still a huge gap between these benchmarks and practical applications.
Specifically, due to the various task budgets in practical application scenes, the number of frames input to the 3D detectors is dynamic, and previous frameworks trained on fixed frame inputs are challenging to generalize.
The cost of training diverse models for tasks with various sizes of inputs is huge, severely limiting the deployment of the models on edge devices.
In order to solve this problem, it is necessary to further research and develop more efficient generalized indoor 3D object detection with variable frames.
For example, in online tasks such as robot navigation~\cite{yin2024sg}, the information captured by the agent is strictly limited by a cost budget constraint, which leads to a varying number of input frames.
As we will show in Section~\ref{bm3}, previous models generalize poorly across different input scales, which brings a huge burden as we need to prepare a series of models trained on different scales of input data to cope with various scenarios.

In practical scenes, the raw input to the indoor 3D object detectors is the RGB-D images with variable frame numbers.
The detectors require the ability to process inputs with various frame numbers at fixed parameters to complete different tasks.
To simulate this, we propose a novel processing framework named AnyView for indoor 3D object detection as shown in Fig.~\ref{teaser}. 
Previous methods require complete scene reconstruction from dense RGB-D images, limiting the model to a fixed number of inputs, which weakens generalization and wastes resources. In contrast, AnyView handles variable frame numbers, making it adaptable to different application budgets. We unify input data as sparse multi-view RGB-D images with camera parameters, compatible with datasets like ScanNet~\cite{dai2017scannet} and SUN-RGBD~\cite{song2015sun}. Instead of concatenating point clouds and extracting fixed global features, AnyView learns geometric clues per frame and uses a transformer-based architecture to efficiently merge and refine these features, enabling processing of variable frame inputs.
We propose a geometric learner to extract local geometric features and a spatial mixtures module to combine these with global semantics, which enables AnyView to generate rich representations from scenes with any number of frames, bridging local geometric and global semantic features.
We introduce randomized view and rectangular dropout strategies to help the model adapt to varying input scales and discard frames with less geometric information. Additionally, a dynamic token strategy ensures consistent feature density and parameter compatibility across point clouds.
Extensive experiments on ScanNet show that our method outperforms previous methods in both accuracy and generalizability with a similar amount of parameters.

\section{Related Work}

\textbf{Point-based 3D object detection:}
This kind of methods take in 3D scenes represented by pure point clouds, which are acquired by LIDAR sensor or 3D reconstruction from multi-view RGB-D images~\cite{shen2023v}. 
Early 3D object detection methods mainly include sliding-window methods~\cite{song2014sliding} and template-based methods~\cite{litany2017asist}. 
Deep learning-based end-to-end 3D object detection methods began to emerge in recent years, which are mainly based on PointNet~\cite{qi2017pointnet} or sparse CNN~\cite{yang2023swin3d, kolodiazhnyi2025unidet3d} backbones. 
PointNet-based methods~\cite{wu2024boosting} consume point clouds directly with the set abstraction operation~\cite{qi2017pointnet++}, which enables flexible receptive fields for 3D feature learning. 
Sparse CNN-based methods~\cite{tong2023multi} project the point clouds to regular grids to be processed by advanced 2D or 3D CNN architectures. 
However, point-based detection models have high inference latency due to inefficient sampling algorithms.

\textbf{RGBD-based 3D object detection:}
RGBD camera is the mainstream 3D sensor for indoor scene understanding tasks. Existing RGBD-based 3D object detection methods take in monocular RGB-D image~\cite{xu2024embodiedsam} or point clouds reconstructed from multi-view RGB-D images (mentioned above). Here we only review the former methods which make full use of both RGB and depth information. 
Prior methods broadly fall into three categories: 2D-driven, 3D-driven, and modal fusion. 2D-driven methods~\cite{xu2018pointfusion} first detect object in images and then use the results to assist search in 3D space.
Kundu \emph{et al.} \cite{kundu20183d} presented a microscopic rendering and comparison loss by employing voxel patterns to represent 3D objects, which allows 3D shapes and poses to be learned via 2D supervision.
However, there is still a huge performance gap between RGBD-based methods and point cloud-based methods.
The most common approach to 3D-driven is to feed image information into a 3D feature extraction network as an additional channel to a point cloud or voxel.
Song \emph{et al.}~\cite{song2014sliding} localized objects on a voxelized point cloud by sliding a 3D detection window, and category information was obtained from RGB pixels attached to the point cloud.
However, this type of simple point cloud and image combination approach may destroy the fine-grained local geometric structural features of the point cloud, resulting in inefficient fusion between the point cloud and image modalities.

\section{Analysis}

\subsection{Problem Statement}\label{bm3}
In order to better utilize 3D detector in practical tasks where the raw input is a variable number of RGB-D images, we propose a new setting for indoor 3D object detection.

\begin{table}[t]
    \caption{Generalizability of previous model. The gray cell indicates on which benchmark the model is trained.}
    \centering
    \setlength\tabcolsep{5pt}
    \vspace{-2mm}
    \footnotesize
    \begin{tabular}{|c|p{1.6cm}<{\centering}|p{1.6cm}<{\centering}|p{1.6cm}<{\centering}|}
        \hline
         & ScanNet-SV & ScanNet-Rec & ScanNet-MV \\
        \hline
        ImVoteNet & \cellcolor{mygray} 31.6 / 13.5 & -- & 43.3 / 20.4 \\
        3DETR & 19.8 / 6.1 & \cellcolor{mygray} 64.5 / 44.0 & 44.7 / 25.6 \\
        \hline
    \end{tabular}
    \vspace{-6mm}
    \label{toy}
\end{table}

\textbf{Inference:}
Unlike the input data format of the existing benchmarks, in order to simulate the inference process of the model in the practical scene, we take the sparse RGB-D images and their camera parameters as the unified input data of the 3D detector to provide the intelligent body with the scene object detection results, which can be summarized as follows:
\begin{equation}
    \mathcal{B}_{N} = \mathcal{D}(\left\{I_{k}, R_{k}^{c}, T_{k}^{c}\right\}_{k=1}^{N}), R_{k}^{c} \in \mathbb{R}^{3 \times 3}, T_{k}^{c} \in \mathbb{R}^{3 \times 1}
\end{equation}
where $N$ represents the number of RGB-D image frames input to the 3D object detector $\mathcal{D}$, and $\mathcal{B}_{N}$ represents the reconstructed point cloud detection results of $N$ frames of RGB-D images.
$I_{k}$, $R_{k}^{c}$, and $T_{k}^{c}$ correspond to the $k_{th}$ frame of RGB-D image, rotation matrix, and translation matrix, respectively.
$\left\{I_{k}, R_{k}^{c}, T_{k}^{c}\right\}_{k=1}^{N}$ can represent both local region and a whole scene. 
To reduce the time for data collection, the maximum number of views is set to 50($N \leq 50$), which is far less than the number required for 3D reconstruction.

\textbf{Model:}
Since the number of input views will change according to the task in practical scene deployments, the 3D object detectors $\mathcal{D}$ must be highly flexible in order to efficiently process RGB-D images with the variable number of frames input.
Therefore, the training process of models $\mathcal{D}$ needs to incorporate adaptations which enable D to process point clouds with arbitrary scale inputs without additional fine-tuning.
This attribute significantly improves the utility and efficiency of the 3D detection applications, enabling them to easily handle a wide range of task input scenes and provide accurate object detection results.

\textbf{Evaluation:}
The performance of model should not be evaluated only on one scale of input. We prepare a variety of validation sets containing different scales of input data, such as monocular input, few-view input and scene-level input. The number of views $N$ can range from $1$ to $50$.
Two representative methods are chosen: ImVoteNet~\cite{qi2020imvotenet} for monocular RGB-D input and 3DETR~\cite{misra2021end} for scene-level point cloud input. 
We prepare three benchmarks from the raw RGB-D videos provided by ScanNet: ScanNet-SV, ScanNet-Rec, and ScanNet-MV\footnote{Refer to Section \ref{bm3yes} for more details.}. ScanNet-SV (single view) is a monocular RGB-D benchmark organized similar to SUN-RGBD.
ScanNet-Rec (reconstruction) is the previous scene-level benchmark. ScanNet-MV (multi-view) shares the same ground truth with ScanNet-Rec but provides multi-view RGB-D images as input instead of reconstructed point clouds.

We train ImVoteNet and 3DETR on ScanNet-SV and ScanNet-Rec respectively and evaluate the models on all benchmarks. When applying 3DETR on ScanNet-MV, we fuse the multi-view depth maps into point clouds according to the camera parameters. As for applying ImVoteNet on ScanNet-MV, we predict bounding boxes for each view and fuse the results by 3D NMS. As shown in Table \ref{toy}, 3DETR trained on reconstructed point clouds performs poorly on the monocular RGB-D benchmark, and both ImVoteNet and 3DETR fail to achieve a satisfactory performance on the multi-view RGB-D benchmark.
This experimental result indicates that the performance of the detector on previous benchmarks may not reflect its performance in practical applications.
And the poor generalizability of previous models poses a huge challenge in applying the existing models in practical tasks, where time for data collection is limited and the scale of input data is changeable.

\section{Approach}

\begin{figure*}[t]
    \centering
    \includegraphics[width=1.0\linewidth]{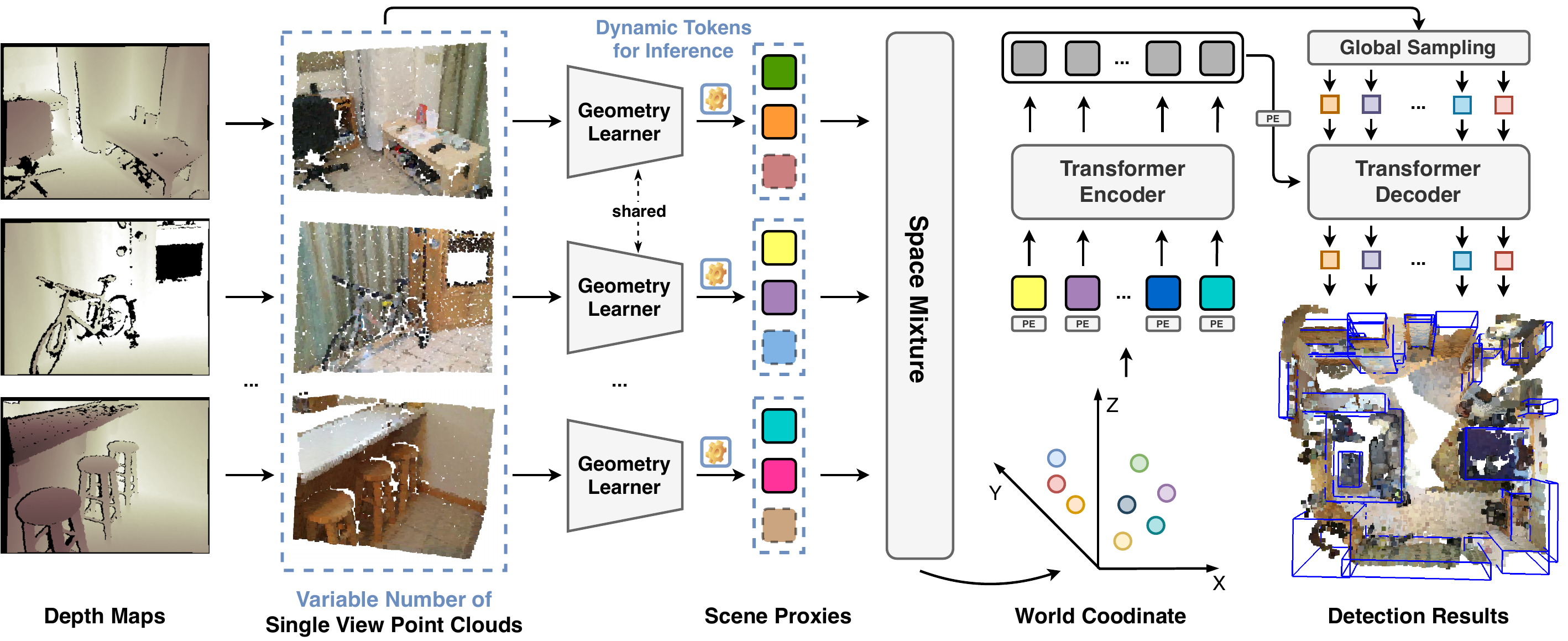}
    \vspace{-6mm}
    \caption{The framework of \emph{AnyView}. Given multi-view RGB-D images of variable number, we first convert the depth maps to single view point clouds according to intrinsic camera parameters. Then a shared geometry learner is applied to extract scene proxies for each view, which are local feature descriptors independent of input scale. Scene proxies from different views are mixed into world coordinates according to the extrinsic camera parameters. We adopt transformer encoder to refine the scene proxies while keeping their input scale-independent property by self-attention mechanism. A transformer decoder is used to refine the object proposals, where the initial proposals are sampled from the whole scene by FPS. The dashed blue box indicates the number of inside elements is changeable during inference.}
    \vspace{-7mm}
    \label{framework}
\end{figure*}

\subsection{Overall Framework}
The overall framework of AnyView is illustrated in Fig. \ref{framework}. 
Given multi-view RGB-D images as input, AnyView focuses on the utilization of depth maps and camera parameters.
We will introduce our method in detail as follows.

Formally, the images are represented by $\{I_1,I_2,...,I_N\}$. We sample $K$ points for each depth map and convert them into camera coordinate by intrinsic camera calibration matrix, which are denoted as $\{P_{c1},P_{c2},...,P_{cN}\}, P_{ci}\in \mathbb{R}^{K\times 3}$. The rotations and translations from camera coordinate to world coordinate for each view are $\{R_{1}^{c},R_{2}^{c},...,R_{N}^{c}\}$ and $\{T_{1}^{c},T_{2}^{c},...,T_{N}^{c}\}$.

\textbf{Extract scene proxies:}
Previous 3D detection methods are only able to process point clouds as a whole. So a natural solution for consuming multi-view RGB-D images is to fuse the depth maps into a whole scene $P=\mathcal{C}(R_{i}^{c}\cdot P_{ci}+T_{i}^{c})$, where $\mathcal{C}$ denotes concatenation operation. 
However, point clouds generated from different numbers of views vary a lot in local geometry structure and global semantics, which will lead to deteriorated scene representation when numbers of view for training and inference are not the same. To decouple scene representation with the scale of input data, we propose to extract $T$ scene proxies for each view independently with a shared geometry learner:
\begin{equation}
    \mathcal{P}_i=G(P_{ci}), \mathcal{P}_i\in \mathbb{R}^{T\times C}
\end{equation}
Scene proxies $\mathcal{P}_i$ are local feature descriptors which represent the geometry structures of single view point clouds. 
We implement the geometry learner $G$ as two set abstraction (SA) layers~\cite{qi2017pointnet++}, the first layer with a constrained receptive field to focus on local geometric details and the second with a large receptive field to aggregate the local details into geometry structures. As $G$ is applied on each view independently, the extracted scene proxies will ignore global semantics and be robust to the scales of input data.

\textbf{Interactions among scene proxies:}
We obtain richer scene representations through interactions among scene proxies. In order to keep the input scale-independent property of scene proxies, we hope the interaction to be linear combination $\mathcal{P}_{ij}=\sum_{i=1}^N\sum_{j=1}^{T}{\alpha_{ij}\mathcal{P}_{ij}}$, where $\sum_i\sum_j{\alpha_{ij}}=1$.
In addition, since the number of view ($N$) is variable, the number of interacting scene proxies is also variable.
Benefiting from the nature of self-attention, these requirements can be elegantly achieved by transformer.

Specifically, we first mix the scene proxies in world coordinates by transforming their coordinates:
\begin{equation}
    \left\{\boldsymbol{P}_{ij}^{x},\boldsymbol{P}_{ij}^{y},\boldsymbol{P}_{ij}^{z}\right\} = R_{i}\cdot \left\{\boldsymbol{P}_{ij}^{x},\boldsymbol{P}_{ij}^{y},\boldsymbol{P}_{ij}^{z}\right\}+T_{i}
\end{equation}
where ${\boldsymbol{P}_{ij}^{x,y,z}}$ represents the spatial position of $\mathcal{P}_{ij}$ in the camera coordinate system, and we project them into the world coordinate system to obtain global information at a higher level with the camera projection matrix.
Then the scene proxies are fed into the transformer encoder with their coordinates converted into Fourier positional embeddings~\cite{tancik2020fourier}. We adopt radius mask~\cite{misra2021end} on the self-attention matrix to conduct interactions from local to global.

\textbf{Decode objects from the whole scene:}
Given the $N_{T} = N\times T$ features from the encoder, we adopt a transformer decoder to refine object queries layer by layer as in DETR~\cite{carion2020end}. The object queries are initial proposals, which are generated by furthest point sampling from the whole scene to ensure coverage:
\begin{equation}
    \mathbf{Q}=\mathcal{M}({\mathcal{F}}({\mathcal{C}}(R_{i}^{c}\cdot P_{ci}+T_{i}^{c})))
\end{equation}
where $\mathcal{M}$ represents the MLP and $\mathcal{F}$ represents the farthest point sampling algorithm.
The refined queries are converted into box parameters and supervised following the procedure in 3DETR.

\subsection{Scale-independent Training}

\textbf{Random view dropping:}
During training, the numbers of views of different inputs in a batch are kept the same in order to parallelize the computation. However, enabling the network to process variable number of views is necessary as it encourages the transformer to adapt to different scale of attention map. To this end, we randomly drop $0$ to $\frac{N}{2}$ views and pad the dropped points with 0. To cut down the interactions between the dropped scene proxies and other tokens (scene proxies for encoder, object queries for decoder), we apply a binary mask on each attention map to indicate which of scene proxies are invalid (with coordinates $(0,0,0)$).
The binary mask will set the inner products between invalid scene proxies and other tokens to $-\infty$, so after softmax operation these attention value will be $0$.

\textbf{Global random cuboid:}
The geometry structure of single view point clouds may vary from view to view. For example, when the distance between the RGB-D camera and the scene varies, the density of point clouds varies as well. When one view only contains floors and walls, the scene proxies extracted from it are uninformative. To allow the network to handle depth maps taken from various shooting situations and learn to ignore scene proxies with less information, we randomly crop a cuboid from the fused point clouds $P={\rm \mathcal{C}}(R_{i}^{c}\cdot P_{ci}+T_{i}^{c})$ and set other points outside the cuboid to $(0,0,0)$.
After that, we transform the point clouds back to the camera coordinates of each view and keep $(0,0,0)$ unchanged.
This strategy makes some views lose geometric information but still provide valid scene proxies, which effectively reduces overfitting and improves the generalizability of AnyView.

\subsection{Dynamic Tokens for Inference}

AnyView extracts $N_{T}$ scene proxies for a $N$-view input scene. In our setting, $N$ may vary from 1 to 50 during inference time, which results in an order of magnitude change in the number of scene proxies. In previous work such as VoteNet~\cite{Qi_2019_ICCV} and 3DETR, the encoder of detector extract 1024 or 2048 seed points for one scene, no matter monocular RGB-D input or scene-level point clouds input. To reach this order of magnitude, $T$ should be around 40 when $N=50$ and be around 2000 when $N=1$. For the former case, AnyView only extracts 40 scene proxies for monocular input, which is not enough to fully represent the scene. While for the latter case, the number of scene proxies will be larger than 10000 when there are more than 10 views. This heavy computation cost is unaffordable in practical applications.

To our best knowledge, all previous networks which use SA layers to extract point features keep a fixed output point number during training and inference. However, as the parameters in SA layer are only relevant with the number of channels, changing the output number seems feasible. The reason why previous methods fix the output number are two-fold: 1) if the output number of a SA layer is changed, the shape of input tensor for the next layer will change and may be incompatible with the parameters; 2) even if the next layer can process input in dynamic shape, the density of point clouds in this level is different from the one during training, which makes the layer unable to extract accurate feature representation.
To overcome this problem in our case, we devise a dynamic token strategy for the geometry learner by keeping the output number of the first SA layer fixed and change the output number of the second one. Thus $T$ can be defined as:
\begin{equation}
    T={\rm min}\{Z / N, O_{SA_{1}}\}
\end{equation}
where $Z$ is a predefined constant and $O_{SA_{1}}$ is the the output number of the first SA layer.
In this way, the second SA layer receives the same number of input points as in training time and thus the output features (i.e.\ scene proxies) are accurate. Although the number of scene proxies varies, the transformer encoder is still able to interact and refine them due to the linear combination nature of self-attention mechanism and our scale-independent training strategy.

Meanwhile, our proposed AnyView can be well applied to the online detection task as it extracts features $\mathcal{P}_{i}$ individually for each view $V_{i}$ when processing RGB-D inputs and then integrates the features of different views $\left\{\mathcal{P}_{1}, ..., \mathcal{P}_{i}\right\}$ in the world coordinate system through an attention mechanism.
The ability to detect objects online fast and accurately in dynamic 3D scenes is crucial for autonomous navigation, environment understanding, and other tasks in the field of autonomous driving and robotics.
Our proposed method provides a reliable and efficient solution for online detection, which can enable robots to effectively perceive and interact with the environment.
  
\section{Experiment}
\begin{table}[]
    \centering
    \setlength\tabcolsep{5pt}
    \caption{Number of parameters of different 3D detectors.}
    \vspace{-2mm}
    \footnotesize
    \begin{tabular}{|p{1.6cm}<{\centering}|p{1.6cm}<{\centering}|p{1.6cm}<{\centering}|p{1.6cm}<{\centering}|}
        \hline
        ImVoteNet & ImVoxelNet & 3DETR-m & AnyView \\
        \hline
        1.8M &104.6M &7.4M &7.6M \\
        \hline
    \end{tabular}
    \vspace{-4mm}
    \label{para}
\end{table}

\subsection{Experiments Setup}

\textbf{Datasets and benchmarks:}\label{bm3yes}
We conduct experiments on the ScanNet~\cite{dai2017scannet} dataset, which is a richly annotated dataset of indoor scenes with 1201 training scenes and 312 validation scenes. 
For each scene, ScanNet provides RGB-D video as well as the reconstructed mesh. We uniformly sample 50 frames from the video for each scene, which serves as the maximum data available for training and evaluating in our proposed setting.
Meanwhile, we further validate that the proposed approach has the ability to accurately detect objects in practical deployment scenes by verifying the performance of each model on the online detection task.
We set three benchmarks as defined in Section \ref{bm3}. ScanNet-Rec is the mainstream scene-level benchmark used by previous works~\cite{hou20193d,Qi_2019_ICCV,misra2021end}, whose input data is reconstructed point cloud of a whole scene and ground-truth is generated by computing the axis-aligned bounding boxes~\cite{hou20193d} for objects in 18 selected categories.
In addition, we designed ScanNet-Online benchmarks to explore the performance of previous 3D object detection models with AnyView in practical deployment scenes. Specifically, the input to 
ScanNet-MV shares the same ground truth with ScanNet-Rec, while the input data are multi-view RGB-D images. If not additionally mentioned, the default number of views is 50. ScanNet-MV is also a scene-level benchmark, which means all RGB-D images from one scene are considered as a single input sample for the network.
ScanNet-Online is each frame in a sequence of RGB-D images, and the model needs to dynamically detect objects in the scene based on the new input frames.
Compared with ScanNet-MV, ScanNet-Online inputs each frame of RGBD images in sequence rather than directly inputting all frames at the scene level, which is more suitable for practical deployment requirements.
ScanNet-SV is a monocular RGB-D benchmark with a single view RGB-D image as input data. For each image, we select the bounding boxes whose center points are within the image from the corresponding scene as its ground truth.

\textbf{Compared methods:}
We train VoteNet~\cite{Qi_2019_ICCV} and ImVoteNet~\cite{qi2020imvotenet} on ScanNet-SV. When applying the models on ScanNet-MV, we predict bounding boxes for each view and fuse the predictions by 3D NMS.
3DETR~\cite{misra2021end} is chosen as the model trained on ScanNet-Rec, which is the mainstream transformer-based 3D detector with less inductive bias.
The comparison between 3DETR and AnyView is particularly relevant since they share similar amount of parameters and both adopt DETR-like detection decoder.
To apply 3DETR on ScanNet-MV, we preprocess the multi-view RGB-D images by fusing the depth maps from each view according to the camera parameters.
We train our AnyView model as well as 3DETR and ImVoxelNet~\cite{rukhovich2022imvoxelnet} on ScanNet-MV, each model representing an input modality: 3DETR for scene-level point clouds, ImVoxelNet for multi-view RGB and AnyView for multi-view RGB-D.
We list the number of parameters of these models in Table \ref{para}.

\begin{table}[t]
    \setlength{\tabcolsep}{5pt}
    \centering
    \small
    \caption{3D object detection results (mAP@0.25 and mAP@0.5) on scene-level and monocular benchmarks. Contents in brackets indicate the method modality, where (D) denotes depth only and (RGBD) means added RGB images.}
    \vspace{-2mm}
      \begin{tabular}{c|l|c|c|c|c}  
        \toprule
        &\multirow{2}{*}{Method} & \textbf{Training} & \textbf{Evaluation} & \multicolumn{2}{c}{mAP}  \\
        && Benchmark & Benchmark &@0.25 &@0.5 \\ 
        \midrule
        \multirow{12}{*}{\rotatebox[origin=c]{90}{Scene-level}} &3DETR & Rec & Rec &62.7 &37.5 \\
        &3DETR-m & Rec & Rec &65.0 &47.0 \\
        \cmidrule(r){2-6}
        &3DETR & MV(D) & Rec &53.6 &33.0 \\
        &3DETR-m & MV(D) & Rec &56.9 &37.8 \\
        \cmidrule(r){2-6}
        &3DETR & Rec & MV(D) &37.8 &22.2 \\
        &3DETR-m & Rec & MV(D) &44.7 &25.6 \\
        \cmidrule(r){2-6}
        &VoteNet & SV(D) & MV(D) &40.9 &20.6 \\
        &ImVoteNet & SV(RGBD) & MV(RGBD) &43.3 &20.4 \\
        \cmidrule(r){2-6}
        &ImVoxelNet & MV(RGB) & MV(RGB) &46.6 &25.2 \\
        &3DETR & MV(D) & MV(D) &51.7 &31.0 \\
        &3DETR-m & MV(D) & MV(D) &54.7 &35.3 \\
        \rowcolor{mygray} &AnyView & MV(D) & MV(D) &60.7 &35.8 \\
        \midrule
        \multirow{8}{*}{\rotatebox[origin=c]{90}{Monocular}} &VoteNet & SV(D) & SV(D) &30.1 &13.9 \\
        &ImVoteNet & SV(RGBD) & SV(RGBD) &31.6 &13.5 \\
        \cmidrule(r){2-6}
        &3DETR & Rec & SV(D) &14.7 &3.9 \\
        &3DETR-m & Rec & SV(D) &19.8 &6.1 \\
        \cmidrule(r){2-6}
        &ImVoxelNet & MV(RGB) & SV(RGB) &21.2 &8.3 \\
        &3DETR & MV(D) & SV(D) &24.7 &10.9 \\
        &3DETR-m & MV(D) & SV(D) &27.7 &12.8 \\
        \rowcolor{mygray} &AnyView & MV(D) & SV(D) &32.7 &15.3 \\
        \bottomrule
      \end{tabular}
      \vspace{-4mm}
    \label{tbl:all}
\end{table}

\textbf{Implementation details:}
We downsample the RGB-D videos from ScanNet to a resolution of $320\times 240$.
Following previous setting~\cite{Qi_2019_ICCV}, we sample 20000 points per scene for ScanNet-SV and 40000 points per scene for ScanNet-Rec as the input point clouds. 
For ScanNet-MV, we sample 20000 points per view for methods that make predictions on each view independently and sample 40000 points per scene for methods that fuse multi-view point clouds as a whole.
While for AnyView, we sample 5000 points per view to reduce computation cost.

The geometry learner of AnyView consists of two SA layers, the first with radius $0.2m$, output number of points 256 and MLP channels $[3,64,128,256]$, the second with radius $0.8m$ and MLP channels $[256,256,256,256]$. During training, we set $T=40$. While in inference time, we set $Z=2000$.
Following the configurations of 3DETR-m, we adopt 3 transformer encoders and 8 decoders in AnyView. The radius mask for encoder is set to $[0.8m,0.8m,1.2m]$.
In terms of data augmentation, both random view dropping and global random cuboid are applied with probability 0.75.

As for the online detection setting, in processing the online detection RGB-D input flow $F=\left\{f_{1},..., f_{i} \right\}$, our proposed AnyView will save the extracted $\mathcal{P}_{i}$ for each RGB-D image frame $f_{i}$, at the input of the next RGB-D frame $f_{i+1}$, it will directly read $\left\{\mathcal{P}_{1}, ..., \mathcal{P}_{i}\right\}$, which will be fed into the detector along with $\mathcal{P}_{i+1}$ to predict the result.

\begin{table*}[]
	\centering
	\setlength{\abovedisplayskip}{0pt}
	\setlength{\belowdisplayskip}{0pt}
    \small
    \caption{3D object detection results (mAP@0.25/0.5) on ScanNet-SV/ScanNet-MV. Uniform/Continuous means the views are sampled uniformly/adjacent from the whole 50 views. Gray cells show on which benchmark the model is trained. Contents in brackets represent training views number.}
    \vspace{-2mm}
	\begin{tabular}{c|l|p{1.6cm}<{\centering}|p{1.6cm}<{\centering}|p{1.6cm}<{\centering}|p{1.6cm}<{\centering}|p{1.6cm}<{\centering}|p{1.6cm}<{\centering}|p{1.6cm}<{\centering}}
		\toprule
		 & \multirow{2}{*}{Methods} & \multirow{2}{*}{ScanNet-SV} &\multicolumn{6}{c}{ScanNet-MV} \\
         & & &\ 5\ &10 &15 &30 &40 &50 \\
		\midrule
        \multirow{7}{*}{\rotatebox[origin=c]{90}{Uniform}} &ImVoteNet &\cellcolor{mygray} 31.6 / 13.5 &30.2 / 13.7 &37.3 / 17.9 &40.3 / 18.5 &43.5 / 20.6 &42.9 / 20.4 &43.3 / 20.4 \\
        &3DETR-m(10) &29.8 / 12.5 &39.7 / 20.0 &\cellcolor{mygray} 49.1 / 28.0 &50.8 / 29.0 &50.3 / 28.9 &49.6 / 28.9 &48.9 / 27.7 \\
        &3DETR-m(30) &29.1 / 13.1 &36.8 / 19.5 &50.1 / 29.6 &53.1 / 31.4 &\cellcolor{mygray} 55.6 / 33.9 &55.7 / 34.3 &55.6 / 34.7 \\
        &3DETR-m(50) &27.7 / 12.8 &34.9 / 17.9 &48.6 / 29.0 &52.5 / 33.3 &55.3 / 36.3 &55.6 / 35.9 &\cellcolor{mygray} 54.7 / 35.3 \\
        &AnyView(10) &31.3 / 13.7 &42.2 / 20.4 &\cellcolor{mygray} 51.3 / 27.2 &52.8 / 29.0 &53.2 / 29.9 &54.9 / 30.4 &54.8 / 30.1 \\
        &AnyView(30) &30.5 / 12.7 &39.8 / 20.1 &52.4 / 29.2 &57.1 / 31.6 &\cellcolor{mygray} 59.1 / 35.2 &59.9 / \textbf{36.8} &60.1 / \textbf{37.4} \\
        &AnyView(50) &\textbf{32.7} / \textbf{15.3} &\textbf{45.0} / \textbf{22.0} &\textbf{53.8} / \textbf{30.5} &\textbf{57.5} / \textbf{33.5} &\textbf{59.5} / \textbf{35.7} &\textbf{60.6} / 35.7 &\cellcolor{mygray} \textbf{60.7} / 35.8 \\
		\midrule
        \multirow{7}{*}{\rotatebox[origin=c]{90}{Continuous}} &ImVoteNet &\cellcolor{mygray} 31.6 / 13.5 &18.1 / 8.8 &25.2 / 13.5 &30.4 / 15.7 &40.5 / 18.4 &43.1 / 20.0 &43.3 / 20.4 \\
        &3DETR-m(10) &29.8 / 12.5 &19.9 / 9.9 &\cellcolor{mygray} 27.7 / 15.3 &34.0 / 18.1 &44.2 / 24.1 &48.7 / 27.8 &48.9 / 27.7 \\
        &3DETR-m(30) &29.1 / 13.1 &18.5 / 9.6 &26.6 / 15.1 &35.4 / 19.7 &\cellcolor{mygray} 48.0 / 29.4 &54.0 / 33.6 &55.6 / 34.7 \\
        &3DETR-m(50) &27.7 / 12.8 &17.9 / 9.7 &26.6 / 15.8 &34.9 / 19.8 &46.9 / 28.8 &52.7 / 33.0 &\cellcolor{mygray} 54.7 / 35.3 \\
        &AnyView(10) &31.3 / 13.7 &20.6 / 10.6 &\cellcolor{mygray} 30.2 / 15.7 &36.0 / 17.7 &47.9 / 26.1 &53.6 / 28.7 &54.8 / 30.1 \\
        &AnyView(30) &30.5 / 12.7 &19.9 / 10.1 &29.9 / 16.5 &36.8 / 21.4 &\cellcolor{mygray} 52.3 / \textbf{30.0} &57.1 / 34.3 &60.1 / \textbf{37.4} \\
        &AnyView(50) &\textbf{32.7} / \textbf{15.3} &\textbf{22.1} / \textbf{11.3} &\textbf{32.9} / \textbf{18.6} &\textbf{40.6} / \textbf{21.8} &\textbf{54.4} / 29.8 &\textbf{58.7} / \textbf{34.4} &\cellcolor{mygray} \textbf{60.7} / 35.8 \\
		\bottomrule
	   \end{tabular}
	\vspace{-4mm}
	\label{tab4}
\end{table*}

\begin{figure*}[t]
    \centering
    \includegraphics[width=1.0\linewidth]{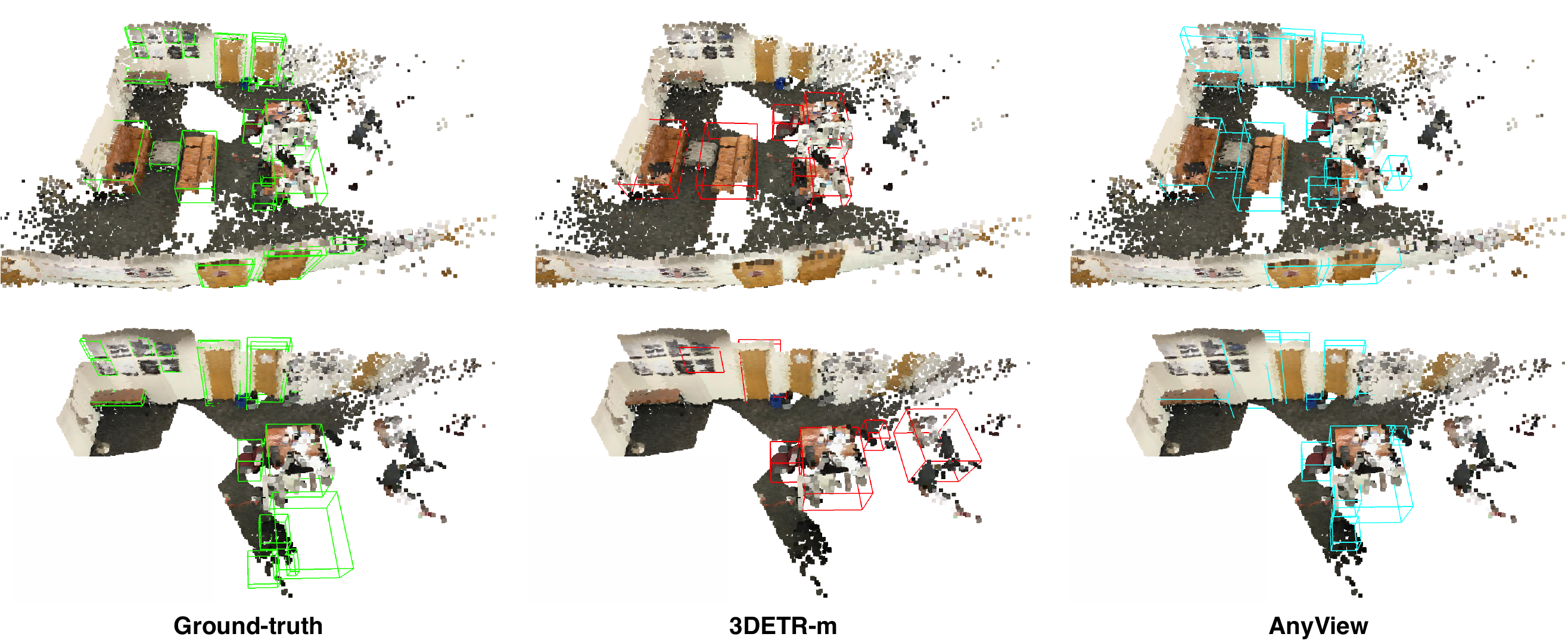}
    \vspace{-8mm}
    \caption{Visual results on ScanNet~(IoU=0.25). We compare the predictions of 3DETR-m and AnyView after NMS with the ground-truth bounding boxes on different scales of input. Top: the whole scene consisting of 50 views, Bottom: a part of scene consisting of 10 adjacent views.}
    \vspace{-6mm}
    \label{vis}
\end{figure*}

\subsection{Results and Analysis}
\textbf{On scene-level and monocular benchmarks:}
We show the performance of different models on scene-level (Rec and MV) and monocular (SV) benchmarks in Table \ref{tbl:all}. As discussed in Section \ref{bm3}, previous models trained on ScanNet-Rec or ScanNet-SV generalizes poorly to other settings. 
Although 3DETR-m trained on ScanNet-Rec achieves the best performance (65.0/47.0) on scene-level benchmarks when evaluated on ScanNet-Rec, it requires reconstructed point clouds which are not available in many practical scenarios. We find 3DETR-m trained on ScanNet-MV gets lower performance (54.7/35.3) on scene-level benchmarks, which indicates the previous ScanNet-Rec benchmark is too idealistic and the point clouds fused from multi-view depth maps are more challenging for 3D object detection.
Among models trained on ScanNet-MV, ImVoxelNet gets relatively lower performance as color is less informative than depth, especially for 3D object detection. 3DETR-m generalizes better to single view inputs (54.7/35.3) when trained on ScanNet-MV, which shows ScanNet-MV is not only a more practical benchmark for evaluation, but also beneficial for training a generalizable 3D detector.
Observing the rows in gray, AnyView achieves leading result (60.7/35.8) on the challenging ScanNet-MV benchmark and also generalizes well (32.7/15.3) to ScanNet-SV, which even outperforms ImVoteNet (31.6/13.5) trained on ScanNet-SV.

\begin{figure*}[t]
    \centering
    \includegraphics[width=1.0\linewidth]{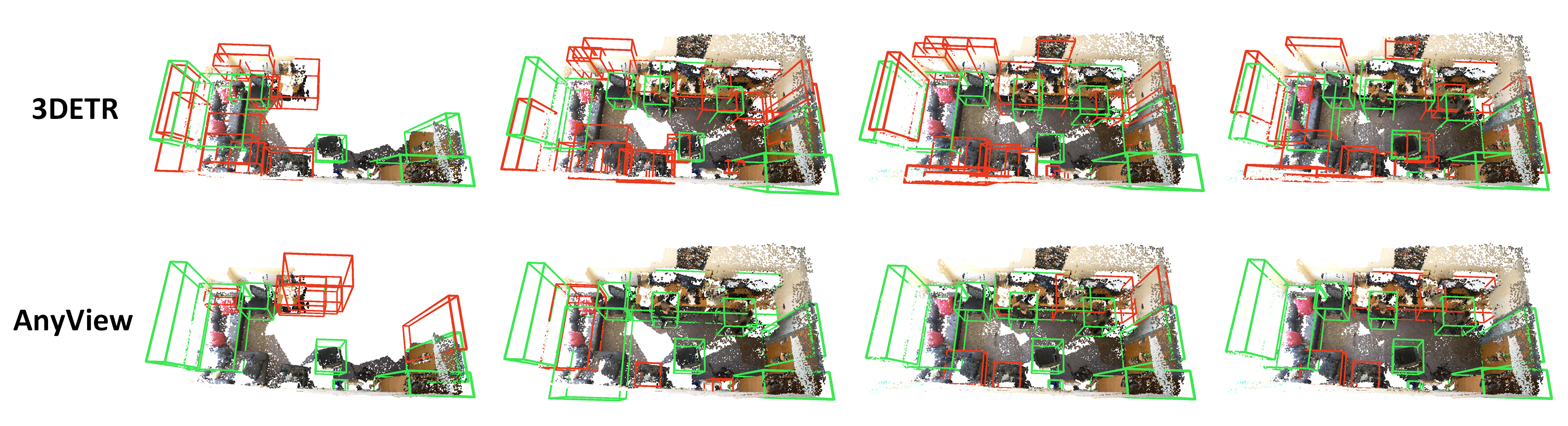}
    \vspace{-8mm}
    \caption{Online detection visualization results~(IoU=0.25). We demonstrate the detection results for 15, 30, 40, and 50 frames during the navigation(scene0474\_00 on the ScanNet dataset).Red bounding boxes and green bounding boxes represent false-positive and true-positive samples, respectively.}
    \vspace{-6mm}
    \label{fig:fig_vis_online}
\end{figure*}

\textbf{On wider range of input views:}
We further extend ScanNet-MV to a series of fine-grained benchmarks for more comprehensive analysis. 
We train models on 10/30/50 views uniformly sampled from the whole 50 views in ScanNet-MV and evaluate them on different numbers of views. We devise two settings for partial scenes: Uniform and Continuous, indicating the evaluation of model is conducted on uniformly/adjacently sampled views.
As shown in Table \ref{tab4}, 3DETR-m trained on small number of views gets low performance on large number of views, and vice versa. On the contrary, with the growing of number of training views, AnyView shows consistent improvement across various scales of input. Even if combining the best results of three 3DETR-m models, AnyView still surpasses them by a large margin with a single set of parameters.

The detection results of 3DETR-m and AnyView after NMS is shown in Fig. \ref{vis}, where a whole scene consisting of 50 views and a part of it consisting 10 adjacent views are chosen as the input.
3DETR-m fails to detect any of the doors for the whole scene. That is because 3DETR-m fuses multi-view point clouds as a whole, which makes the detector hard to focus on thin objects like doors and windows whose geometric information is weak.
It also aggregates outliers into misleading clusters, resulting in false positive.
On the contrary, AnyView extracts scene proxies for each view independently, which owns better understanding of local geometry structures and successfully detects all the doors for both scenes. Benefiting from the input scale-independent property, AnyView outputs consistent predictions for different input scales.

\textbf{Online 3D object detection:}
We conduct online object detection experiments to demonstrate the performance of AnyView with any-scale point cloud input as shown in Fig. \ref{fig:fig_vis_online}. 
The views of each scene are fed into AnyView in a frame-by-frame format, and the aggregated point cloud reconstruction results are fed into the 3DETR.

\begin{figure}[t]
    \centering
    \includegraphics[width=1.0\linewidth]{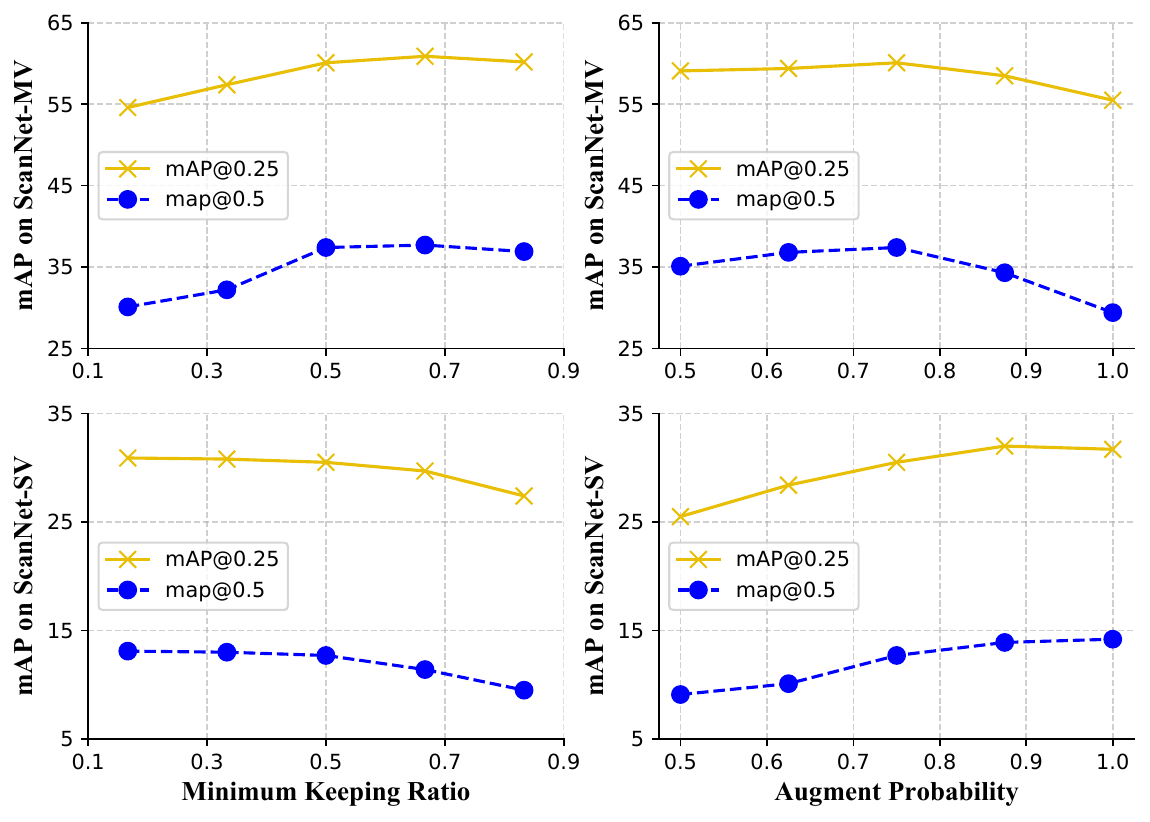}
    \vspace{-8mm}
    \caption{The effects of minimum keeping ratio for random view dropping and probability for data augmentation on the final performance of AnyView on ScanNet-MV and ScanNet-SV.}
    \vspace{-8mm}
    \label{ab2}
\end{figure}

\subsection{Ablation Study}
We conduct ablation studies to show how different architecture designs and training/inference strategies influence the performance of the proposed AnyView framework. The models are trained on ScanNet-MV with 30 views. With a high degree of site cloud fragmentation (15 views), 3DETR exhibits more false positives, while AnyView maintains high detection accuracy. As the number of viewpoints increases, 3DETR’s false positives are somewhat reduced, achieving better results at 50 views.

\textbf{Performance w.r.t. architecture design:}
We ablate three parts of architecture designs of AnyView, as shown in Table \ref{ab1}. GlobalQuery indicates the object queries are sampled from the concatenated scene (\checkmark) instead of the coordinates of scene proxies. CoordsTrans means whether to apply geometry learner in the camera coordinate (\checkmark) of each view or in the world coordinate. PE$_{enc}$ indicates whether to use positional embeddings for the transformer encoder. The first two rows show that high object query coverage of the scene is beneficial for the transformer decoder. The third row shows unifying the coordinate makes the geometry learner extract more robust scene proxies. As the features of point clouds already contain positional information, 3DETR finds positional embeddings are not necessary. However, in AnyView the scene proxies are independently extracted, so the spatial relationship between scene proxies from different views is weakened. We find Fourier positional embeddings with a MLP performs best.

\begin{table}[t]
    \setlength{\tabcolsep}{5pt}
	\centering
    \small
	\caption{The effects of global query sampling, coordinate transformation and encoder positional embeddings on the AnyView with ScanNet-MV.}
    \vspace{-2mm}
	\begin{tabular}{ccc|cc}  
        \toprule
        \multirow{2}{*}{GlobalQuery} &\multirow{2}{*}{CoordsTrans} &\multirow{2}{*}{PE$_{enc}$} &\multicolumn{2}{c}{mAP}   \\
        & & & @0.25 & @0.5 \\
        \midrule
         & & &58.5 &33.1 \\
        \checkmark & & &59.2 &34.4 \\
        \checkmark &\checkmark & &59.6 &34.9 \\
        \checkmark &\checkmark &Fourier &60.6 &35.8 \\
        \rowcolor{mygray} \checkmark &\checkmark &MLP$\circ$Fourier &60.1 &37.4 \\
        \bottomrule
      \end{tabular}
	\vspace{-2mm}
	\label{ab1}
\end{table}

\textbf{Performance w.r.t. data augmentations:}
We further investigate the effects of two proposed data augment strategies. 
ScanNet-MV measures the detection accuracy and ScanNet-SV measures the generalizability of the detector. 
As shown in Fig. \ref{ab2}, with the growing of minimum keeping ratio for random view dropping, the performance of AnyView on ScanNet-MV grows but its generalizability to single view input has deteriorated.
In terms of the probability of augmentation, it is shown that low probability leads to overfitting and poor generalizability, while very high probability hurts the performance of AnyView. 
Therefore we choose dropping $[0,\frac{N}{2}]$ views and augmenting with probability 0.75.

\begin{figure}[t]
    \centering
    \includegraphics[width=1.0\linewidth]{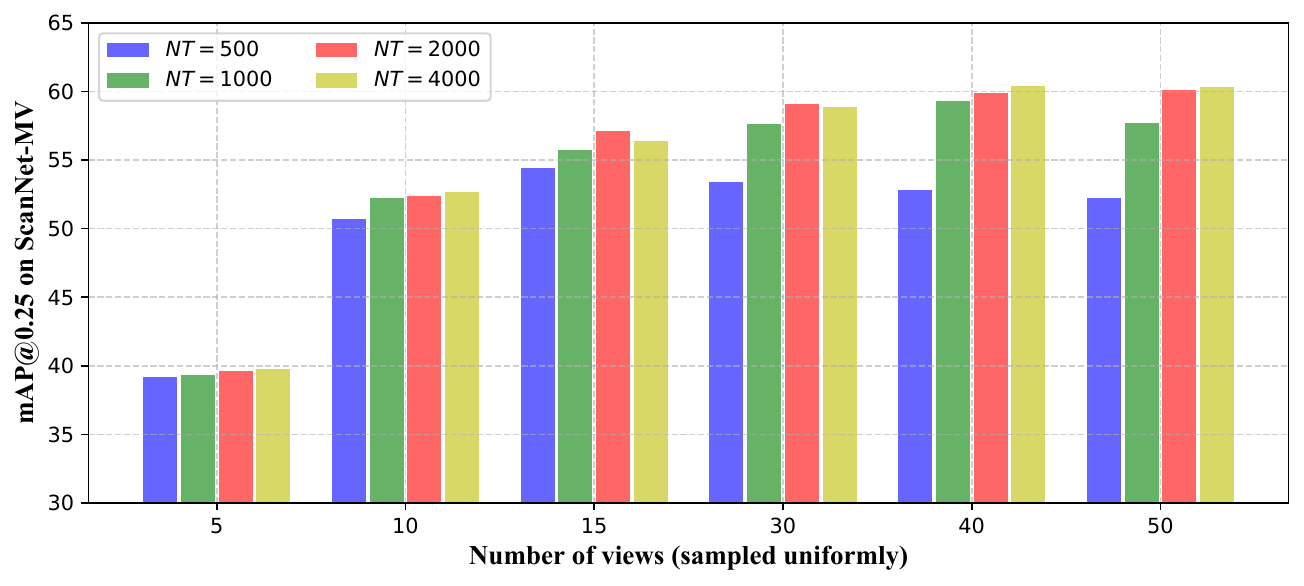}
    \vspace{-8mm}
    \caption{The effects of token number during inference on the final performance of AnyView on ScanNet-MV.}
    \vspace{-6mm}
    \label{ab3}
\end{figure}

\textbf{Performance w.r.t. number of tokens:}
Fig. \ref{ab3} illustrates the effects of our dynamic token strategy, where the number of scene proxies for each view is $T={\rm min}\{X/N, 256\}, X\in \{500,1000,2000,4000\}$. 
We find when the number of scene proxies is too small ($N_{T}\leq 1000$), the performance of AnyView will even drop with the growth of view numbers. 
With the increasing number of $N_{T}$, the performance of AnyView is significantly improved for any view number. 
However, when NT is greater than 2000, the rate of performance increase decreases significantly, introducing additional computational cost without sufficient performance improvement, and $N_{T}=4000$ even leads to a decrease in accuracy.
So finally we set $N_{T}=2000$ for a better trade-off between performance and cost.

\section{Conclusion}
In this paper, we challenge the existing benchmarks for indoor 3D object detection in a perspective of input data.
We propose a new practical setting for this task, which unifies the input modality as multi-view RGB-D images with variable input frame numbers and evaluates 3D detectors on various scales of input data. 
We design a new transformer-based framework named AnyView for practical applications, which is able to process scenes consisting of any number of frames and extract input scale-independent scene representations. 
Benefiting from the nature of the self-attention mechanism and our scale-independent training strategy, AnyView is able to change the number of scene representations extracted for each input frame during inference and flexibly handle various numbers of input frames.
Extensive experiments show that AnyView achieves both great generalizability and high detection accuracy.

\section*{Acknowledgments}
This work was supported in part by the National Natural Science Foundation of China under Grant 62376032, Grant 62125603 and A*STAR National Robotics Programme (RIE2025) M25N4N2009.

{\small
\bibliographystyle{ieee_fullname}
\bibliography{root}
}

\end{document}